\documentclass[10pt,letterpaper]{article}
\usepackage{cogsci}
\usepackage{booktabs}
\usepackage[bottom]{footmisc}
\usepackage{hyperref}
\usepackage{graphicx}
\usepackage{epstopdf}
\usepackage{wrapfig}
\cogscifinalcopy % Uncomment this line for the final submission 
\usepackage[utf8]{inputenc}
\usepackage[
    backend=biber,
    style=apa,
    natbib=true,
    doi=false,
    isbn=false,
    url=false,
]{biblatex}
\usepackage{xcolor}

\usepackage{pslatex}
\usepackage{float} % Roger Levy added this and changed figure/table
\usepackage{amsmath}
\title{AI-enhanced semantic feature norms for 786 concepts}

\author{{\large \bf Siddharth Suresh*$^{1,2,3}$}, 
{\large \bf  Kushin Mukherjee*$^{4}$},
{\large \bf Tyler Giallanza$^{5}$},
{\large \bf Xizheng Yu$^{6}$},
{\large \bf Mia Patil$^{2}$},\\
{\large \bf Jonathan D. Cohen$^{5,7}$,} and
{\large \bf Timothy T. Rogers$^{1,3}$}\\
$^{1}$Dept. of Psychology, University of Wisconsin-Madison,
$^{2}$Dept. of Computer Sciences, University of Wisconsin-Madison, \\
$^{3}$Wisconsin Institute for Discovery,
$^{4}$Dept. of Psychology, Stanford University,
$^{5}$Dept. of Psychology, Princeton University,\\
$^{6}$Dept. of Computer Science, Brown University,$^{7}$Princeton Neuroscience Institute\\
}
\begin{document}

\maketitle

\begin{abstract}

Semantic feature norms have been foundational in the study of human conceptual knowledge, yet traditional methods face trade-offs between concept/feature coverage and verifiability of quality due to the labor-intensive nature of norming studies. 
Here, we introduce a novel approach that augments a dataset of human-generated feature norms with responses from large language models (LLMs) while verifying the quality of norms against reliable human judgments.
We find that our AI-enhanced feature norm dataset, \texttt{NOVA}: \textbf{N}orms \textbf{O}ptimized \textbf{V}ia \textbf{A}I, shows much higher feature density and overlap among concepts while outperforming a comparable human-only norm dataset and word-embedding models in predicting people's semantic similarity judgments.
Taken together, we demonstrate that human conceptual knowledge is richer than captured in previous norm datasets and show that, with proper validation, LLMs can serve as powerful tools for cognitive science research.

\textbf{Keywords:} 
semantic knowledge; feature listing; large language models; similarity judgments
\end{abstract}

\section{Introduction}

The study of human conceptual knowledge has relied on semantic feature norms --- representations of concepts in terms of their associated features --- since their introduction by Rosch in the 1970s \citep{rosch1975cognitive}.
Norming studies present participants with a set of concepts and, for each, asks them to list as many characteristic properties as they can. Aggregating features across items and participants creates semantic vectors the elements of which correspond to the elicited features and the entries of which indicate whether people regularly judge the concept to possess the corresponding property. Proximity between two such feature vectors relates systematically to their perceived semantic relatedness---thus lions and tigers are viewed as similar kinds of things because they have many overlapping and few distinguishing properties. Norming datasets collected over the years \citep{mcrae2005semantic, devereux2014centre, buchanan2019english, ruts2004dutch, hansen2022semantic,dilkina2008semantic} have helped to answer questions about the organization of semantic memory \citep{collins1975spreading,ashcraft1978property}, its degradation in semantic disorders \citep{farahmcclelland1995,rogersETAL2004, garrard2001longitudinal,creemcrae03}, its relationship to control \citep{giallanzaETAL2024}, and its neural bases \citep{CoxETAL2024,clarke2014object} (see \textcite{kumar2021semantic} for a review).

Semantic norming requires extensive human labor both in data collection and curation/post-processing. Prior studies have met this challenge in different ways, each requiring some degree of compromise as elaborated below. Other recent work has sought alternatives to human feature norms by making use of natural language processing technologies, including word embeddings from methods such as \texttt{word2vec} and \texttt{GloVe}  \citep{pennington2014glove, mikolov2013distributed} as well as feature norms generated artificially by large language models (LLMs) \citep{hansen2022semantic}. 
However, word embeddings fail to capture the semantic structure perceived by humans as effectively as feature norms, and their dimensions lack the transparent interpretability of feature-based representations, at least for for concrete objects \citep{suresh2023conceptual,suresh2023semantic}. 
LLMs can generate super-human lists of features that go far beyond what a typical person might know (and hence are non-representative of human knowledge) and frequently confabulate properties that are untrue (the well-documented `hallucination problem' \citep{huang2024survey}).

The current work seeks a middle way between human-only and machine-only norm generation. We crowd-sourced feature lists for a modestly large and representative set of 786 concrete object concepts thus ensuring that the features included in the set are those that human participants discern. We then used LLMs to aid in the most labor-intensive parts of data curation and post-processing, resulting in a novel {\em AI-enhanced} set of semantic feature norms -- \texttt{NOVA}: \textbf{N}orms \textbf{O}ptimized \textbf{V}ia \textbf{A}I. 
We illustrate remarkable differences between human-only and AI-enhanced norm sets, then report empirical studies designed to assess whether the AI-enhanced norms capture human-perceived semantic structure better than do human-only norms or ``out-of-the-box'' word embeddings.

\section{Study I: Building NOVA}

\begin{figure*}[ht!]
\centering
\includegraphics[width=.75\textwidth]{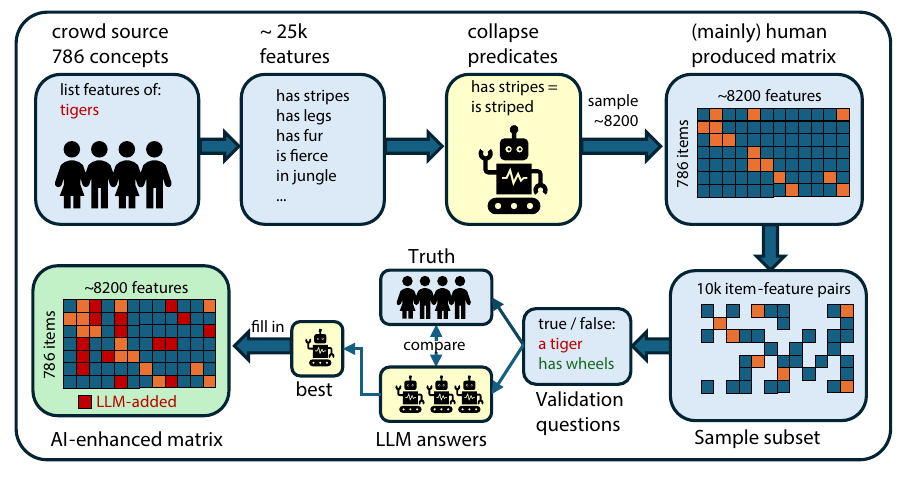}
\caption{A schematic representation of our workflow. Features were initially crowd-sourced for 786 concepts, forming a human-generated matrix. A subset of 10,000 concept-feature pairs underwent validation via human judgments. LLM responses were compared to these human judgments to determine the best-performing strategy. Using this method, LLMs completed the matrix for all 8,200 selected features, forming the AI-augmented matrix.
}
% \vspace{}
\label{fig:schematic}
\end{figure*}

% \subsection{Overview of the approach}
Human feature-norming studies involve up to 4 steps, each requiring significant effort and thus subject to constraints that can limit the resulting data. Here we consider each step, limitations faced by prior studies, and the approach taken in the current work. The overall workflow for our approach is shown in Figure~\ref{fig:schematic}.

\noindent{\em Concept selection.} 
The structure appearing in a given dataset depends on the concepts included. 
Early norms used in semantic memory studies focused on hierarchically structured, easily nameable concepts (e.g., animals, plants), often excluding typical examples (e.g., robins, sparrows) in favor of atypical ones easier to name (e.g., penguins, ostriches) and omitting concepts that don't fit neatly into these hierarchies. 
To improve representativeness we included all 565 concepts from the Ecoset dataset \citep{mehrer2021ecologically}, which comprises frequent, unambiguous basic-level concrete object names, along with items from the McRae \citep{mcrae2005semantic} and Leuven \citep{leuven} norms. We also added superordinate categories (e.g., animal, vehicle) and higher-frequency subordinate names (e.g., robin, trout) to better capture domain substructure. The final set comprised \textbf{786} concrete object concepts.

\begin{figure*}[ht!]
\centering
\includegraphics[width=.75\textwidth]{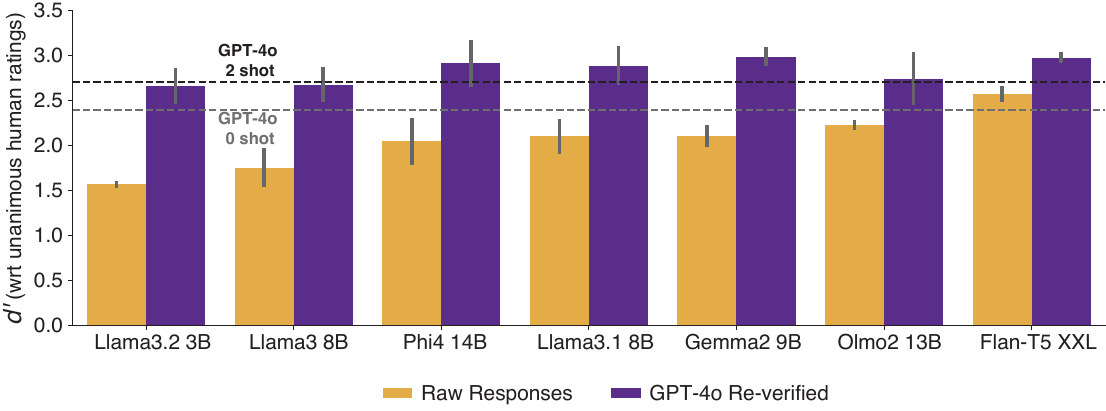}
\caption{Models' ability to reliably predict human feature-concept ratings measured as $d'$ using raw responses (orange) and responses re-verified using GPT-4o. Bar heights show mean $d'$ across the 0-shot and 2-shot experiments.
Gray and black dashed lines correspond to GPT-4o's performance in the 0-shot and 2-shot settings respectively.
Errorbars correspond to bootstrapped 95\% confidence intervals.
}
\vspace{-5pt}
\label{fig:model_evals}
\end{figure*}

\noindent{\em Feature elicitation.} We elicited features from participants on Amazon Mechanical Turk using procedures described below.

\noindent{\em Feature reduction.} Norming studies typically yield a large set of unique features, most appearing in a single concept. To manage this complexity, researchers often consolidate distinct yet semantically related properties -- e.g., if \textit{is hairy} and \textit{is furry} are used by different participants to describe a `coconut', these features may be deemed as equivalent, creating a single feature that overlaps for concepts possessing both \textit{is hairy} (e.g., ape) and \textit{is furry} (e.g., rabbit). 
While this process simplifies the feature space and enhances conceptual similarity across items, it is labor-intensive and relies on subjective human judgments. We instead performed a minimal feature collapse by using GPT-3 to extract phrase embeddings of featural descriptions  (e.g., \textit{has a furry outer layer}), then clustering these and merging only highly similar clusters. This approach collapsed phrases with variable wording but near-identical semantic content (e.g., \textit{has a furry outer layer`},\textit{is furry}, and \textit{feels furry)}) while still distinguishing close synonyms (e.g., \textit{is hairy} vs \textit{is furry}).
This step reduced the initial $\sim$25k raw features to $\sim$20k features, from which we randomly sampled $\sim$8,200 features for subsequent analysis.

\noindent{\em Feature verification.} The features that participants generate in the elicitation phase typically constitute a fraction of what they actually know. For this reason, some norming studies conduct a {\em feature verification} step where human participants consider every concept/feature pair and judge whether the feature is true of the concept \citep{leuven,dilkina2008semantic}. This step greatly enriches the structure encoded in the norms. For instance, most participants list the feature\textit{has a long neck} for giraffes and swans but for few other items. Yet when asked, most participants agree that \textit{has a long neck} is true of items as varied as a duck, a beer bottle, and a cello. Thus, the verification phase surfaces latent knowledge that participants don't generate spontaneously. Since the number of concept/feature pairs grows exponentially, this is by far the most labor-intensive part of the process and prior studies have either employed a relatively modest set of concepts and features \citep{dilkina2008semantic} or have limited verification only to specific semantic domains \citep{leuven}. 
We leveraged LLMs to conduct the feature-verification phase -- first comparing different models and strategies in their ability to capture human judgments on a randomly-sampled set of concept-feature pairs, then using the most successful strategy to verify all $\sim$6.5M concept/feature pairs, producing an AI-enhanced norm set.

\begin{figure}[b!]
    \centering
    \includegraphics[width=.85\linewidth]{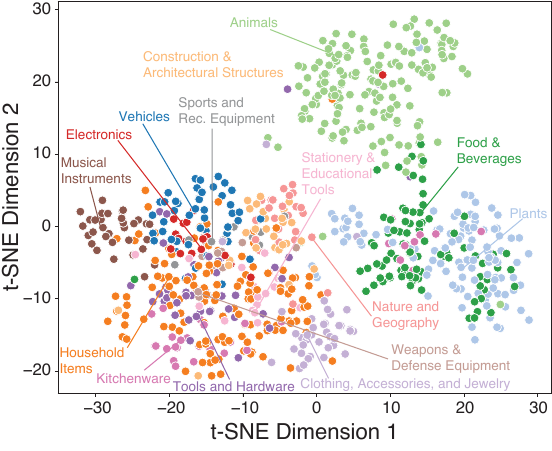}
    \caption{$t$-stochastic neighbor embeddings of the semantic vectors for each of 786 concepts derived from the final verified matrix. Category labels were generated by combining higher order labels from existing norm datasets and LLM-suggested categories from GPT-4o.}
    \label{fig:embeddings}
    % \vspace{-10pt}
\end{figure}

\subsection{Methods}
\textbf{Human feature elicitation.}
This phase provided human-elicited data for all concepts in the set, providing the raw features from which human-only and AI-enhanced norms were derived.

\noindent\textit{Participants.}
50 participants were recruited through Amazon Mechanical Turk and were compensated \$4 for the task which would require 20 minutes to complete.
The study was approved by the Princeton Internal Review Board, IRB Protocol 6079.

\noindent\textit{Stimuli and procedure.}
Stimuli were 786 concrete object nouns. Using a web-based interface, each participant viewed up to 75 different words in randomized order, and for each typed in as many different features as they could generate. The instructions emphasized generating various types of features, including physical/perceptual features (appearance, smell), functional features (uses, contexts), and other characteristics. Participants were asked to format their responses as individual features per line using standardized phrasing (e.g., ``has ears'' rather than ``a dog is an animal that has ears'').

\subsubsection{Human feature verification.}
This phase had human participants verify $\sim$10k concept-feature pairs, providing an empirical basis for evaluating the performance of different AI-aided approaches to feature verification.

\noindent\textit{Participants.}
556 participants were recruited through Amazon Mechanical Turk and compensated \$1.40 for a 5-8 minute task. Participants were allowed to complete multiple sessions contingent upon maintaining satisfactory performance.

\noindent\textit{Stimuli and procedure.}
The stimuli were concept-property pairs sampled randomly from results of the feature-elicitation task. Data were collected through an online interface. Each trial paired one concept (e.g. ``alligator'') with one feature randomly sampled from the full set. The sampled feature could come from any domain or item--for alligator, it could be something reasonable (e.g. ``has legs''), something clearly false (e.g. ``has wheels'') or something uncertain (e.g. ``has ears''). For each pair participants judged whether the property is true of the item by pressing a keyboard button. The instructions emphasized that subjective properties should be evaluated based on common consensus (e.g., ``cute'' for ``dog''), and properties that were sometimes true should be marked as true (e.g., ``brown'' for ``dog''). Each participant made about 110 judgments, and we collected 5 or more judgments on each of 10,545 unique pairs. Participants could skip unfamiliar concepts or nonsensical properties by pressing the space bar, with skipped items replaced to maintain the required number of judgments.
\begin{figure*}[t!]
    \centering
    \includegraphics[width=0.75\textwidth]{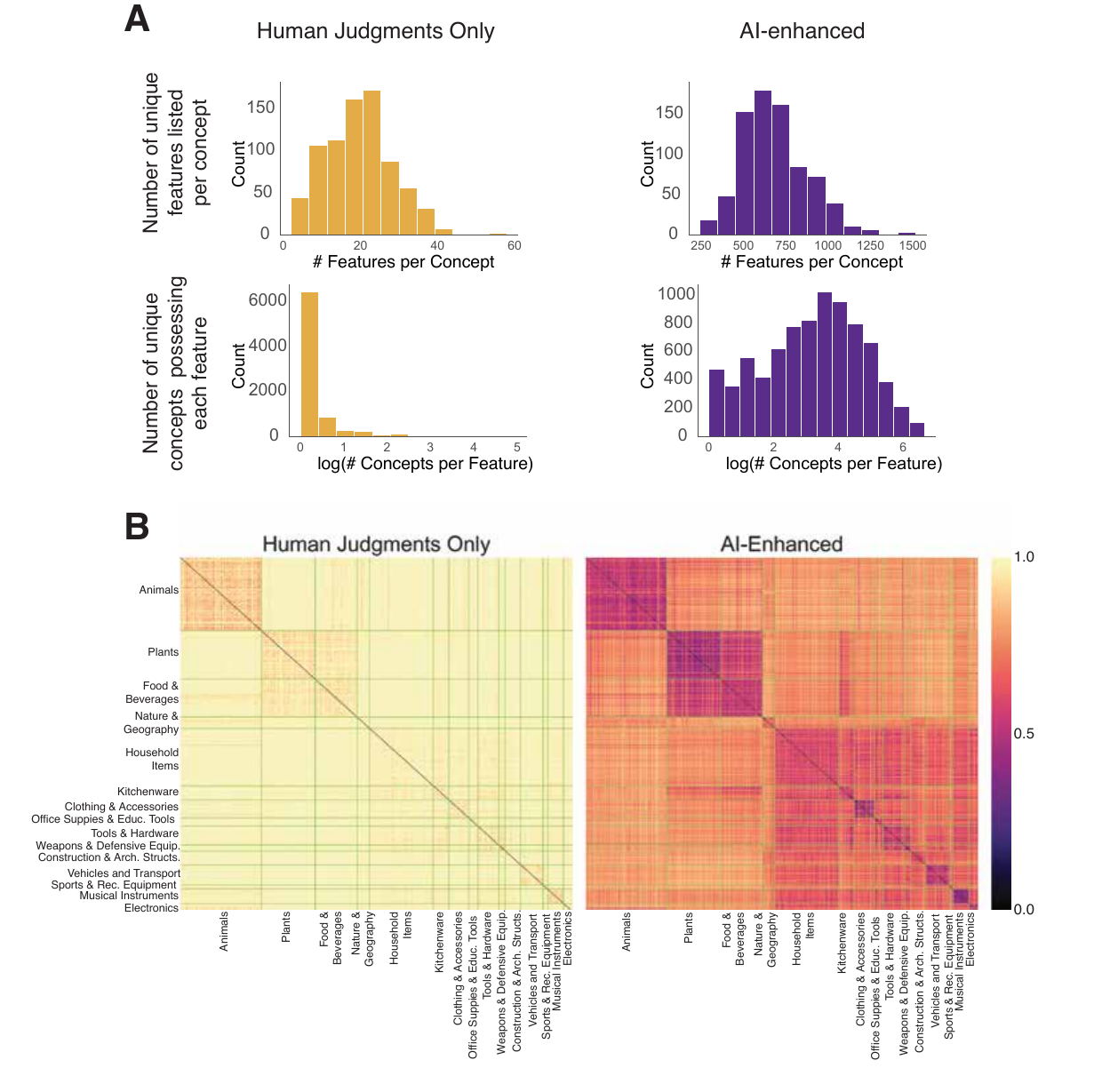}
    \caption{(A) Counts of valid features per concept and number of concepts that share common features for the reduced human-generated matrix (top row) and  AI-enhanced norm matrix (bottom row). (B)  Pairwise cosine dissimilarity matrices based on the reduced human-generated norm matrix (left) and AI-enhanced norm matrix (right).  }
    \label{fig:heatmap_stats}
    \vspace{-10pt}
\end{figure*}

\textbf{AI-enhanced feature verification.}

Our ultimate goal was to use LLMs to complete the feature-verification step for all possible concept/property pairs. Since there are millions of possible pairs, we first considered how well each of several different models and prompting strategies could capture real human judgments on the items collected in the human feature-verification study. In these data participants showed different opinions for about 40\% of the items--thus either opinion expressed by an LLM would agree with at least one human participant for these items. We therefore selected the 6,122 concept-feature pairs for which all participants made the same decision (either all yes or all no), and used these decisions as a ground-truth for evaluating LLM performance.

\noindent\textit{Model Suite.} We primarily focused on performant open-sourced language models because these are accessible to other researchers for replication purposes and relatively more affordable to access. We included models that have open weights, are generally high-scoring on standard LLM benchmarks \citep{hendrycks2020measuring}, and can be run on consumer-grade hardware. Specifically, we evaluated 3 models from Meta's \texttt{Llama} family (\texttt{Llama3}, \texttt{Llama3.1}, and \texttt{Llama3.2}) \citep{dubey2024llama}, Microsoft's \texttt{Phi-4} \citep{abdin2024phi}, Ai2's \texttt{Olmo2} \citep{olmo20242}, and Google's \texttt{Gemma2} \citep{team2024gemma} and \texttt{Flan-T5} \citep{flan}.
We evaluated all models at full \texttt{bfloat16} precision on a Nvidia H100 GPU. For comparison to a state-of-the-art closed model, we also evaluated GPT-4o via its API.

\noindent\textit{Evaluation Protocol.}
We prompted all models using the following general prompt - 
\begin{verbatim}
In one word True or False, answer the following 
question question: Is the property [x] true for
[y]? Answer:
\end{verbatim}
...where $x$ was a feature and $y$ was a concept with the square brackets included in the prompt.
We ran two prompting experiments: (1) a zero-shot experiment providing the models with just the question above as input, and (2) a two-shot experiment providing the models with two example feature-concept pairs, one true and one false, to potentially improve the models' ability to perform the task via in-context learning \citep{brown2020language}. We used the same two examples for all prompts.
%\km{contemplating whether to include those here... TBD}

\noindent\textit{Post-processing.} To extract meaningful answers from model-generated text we first restricted responses to a maximum of five tokens, then conducted a case-insensitive search of model responses for the strings `True' or `Yes' to indicate a positive response, and `False' or `No' to indicate a negative response. In rare cases where no match was found we set the model response to `False'.

\subsubsection{Results.}

To measure how closely LLM responses aligned with unanimous human judgments for the 6,122 feature-concept pairs, we adopted a signal detection approach, treating human responses as the true signal and model responses as guesses. Where humans agreed the property was true of the concept, model guesses were scored as hits if they concurred and misses otherwise. Where humans agreed the property was not true of the concept, model guesses were scored as correct rejections if they concurred and false alarms otherwise. From these counts we computed hit rates and false alarms rates, then converted these to the $d'$ measure of signal discrimination. 

The average $d'$ for both zero and two shot conditions can be seen in Figure \ref{fig:model_evals} (yellow bars). 
Two-shot \texttt{GPT-4o} outperformed all open-sourced models, which varied in their match to human responses. Two-shot \texttt{Flan-T5 XXL} performed best amongst open-sourced models and better than the zero-shot \texttt{GPT-4o}. \texttt{Flan-T5's} lower $d'$ relative to \texttt{GPT-4o} was driven by a propensity to respond with `true' to many queries, buoying its hit rate but also increasing its false-positive rate. 
To preserve the benefits of \texttt{GPT-4o} without incurring a prohibitive cost, we next considered a `re-verification' approach in which the `true' responses generated by a given open-source model were subsequently re-verified by \texttt{GPT-4o}, retaining the `true' value only if both models agreed. The results are shown as purple bars in Figure \ref{fig:model_evals}. Re-verification improved performance for all models, surpassing \texttt{GPT-4o} alone. \texttt{Flan-T5 XXL} remained a top model, closely matched by \texttt{Gemma2 9B}. Given the strong baseline performance of \texttt{Flan T5}, we chose this model with \texttt{GPT-4o} re-verification to fill out the full semantic feature matrix.

\begin{figure}[t!]
    \centering
    \includegraphics[width=.76\linewidth]{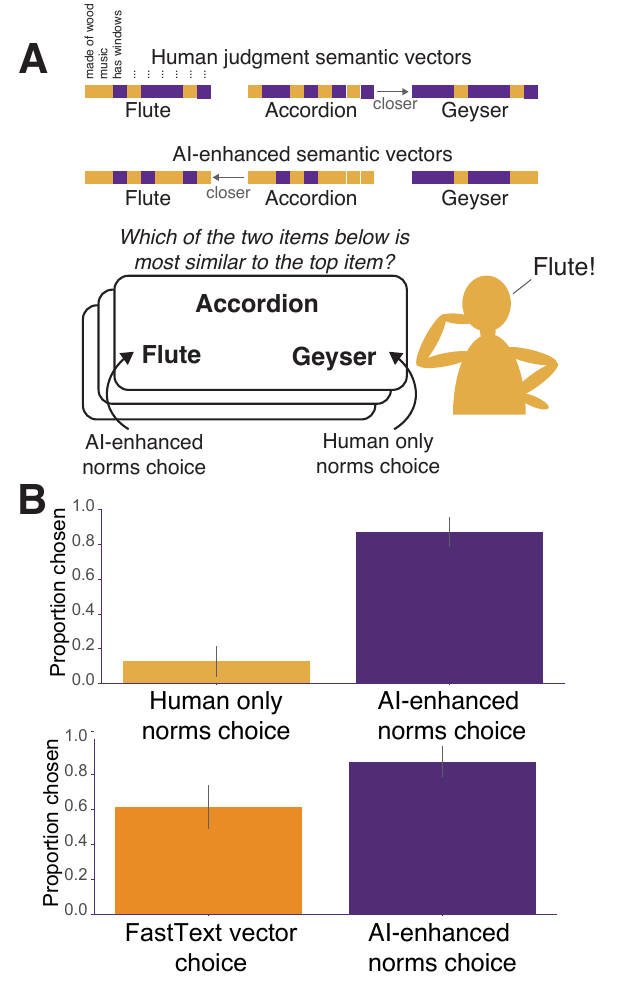}
    \caption{(A) Procedure for generating trials for the triadic judgment experiment and an example trial. (B) Proportion of human responses that aligned with the human matrix (yellow bar) vs. the AI-enhanced matrix (purple bar) and with FastText word embeddings (orange) vs. AI-enhanced semantic vectors (purple) in Experiment 2. Error bars represent standard errors of the means.}
    \label{fig:part2_results}
    \vspace{-15pt}
\end{figure}

\noindent\textbf{Using Flan T5 and GPT-4o to impute the AI-enhanced matrix.} In the human-only matrix, entry $[i,j]$ has a value of 1 wherever a participant produced feature $j$ for concept $i$ and a 0 in all other entries. For every 0 in this matrix, we prompted \texttt{Flan T5 XXL} to decide whether the corresponding property is / is not true of the corresponding concept. Where the model decided `not true,' the zero value was retained in the matrix. Where the model decided `true,' (534,010 out of 6,436,554 possible pairs) we prompted \texttt{GPT-4o} with the same pair to re-verify the answer. If \texttt{GPT-4o} agreed the property was true, the cell value was replaced with 1, otherwise the 0 value was retained. This procedure yielded the final \textbf{AI-enhanced} norms matrix. Figure \ref{fig:embeddings} shows t-SNE based embeddings of all concepts from this matrix.

The AI-enhanced matrix differed remarkably from the human-only matrix in its feature density. While the human matrix has about 20 features per concept on average, the AI-enhanced matrix has about {\em 700} (Figure \ref{fig:heatmap_stats}A), and while the majority (78\%) of features in the human-only matrix are true of just one concept, this is true of just 5\% of features in the AI-enhanced matrix. The increased feature density produces much more richly-structured similarity relations, as shown by the heat plot of pairwise distances between concepts in Figure \ref{fig:heatmap_stats}B. While some of this difference may be attributable to false-positives in the AI-enhanced dataset, the comparison to human judgments suggests that the LLM verification strategy is quite good at discriminating true positives from true negatives ($d' > 3.0$). Thus the result suggests that human knowledge about features of concepts may be considerably richer than prior norming studies have suggested.

% filepath: table.tex
% \begin{table}[h]
%     \centering
%     \begin{tabular}{llcc}
%     \toprule
%     & Model & 0-shot & 2-shot \\
%     \midrule
%     Raw & \texttt{Llama3.2-8B} & 1.59 & 1.55 \\
%     Responses &  \texttt{Llama3-8B} & 1.95 & 1.56 \\
%     &  \texttt{Phi4-14B} & 2.29 & 1.81 \\
%     & \texttt{Llama3.1-8B}  & 2.28 & 1.93 \\
%     & \texttt{Gemma2-9B}  & 2.21 & 2.00 \\
%     & \texttt{Olmo2-13B} & 2.27 & 2.20 \\
%     & \texttt{FLAN-T5-XXL} & 2.50 & 2.65 \\
%     & \texttt{GPT-4o} & 2.40 & 2.71 \\
%     \midrule
%     \midrule
%     GPT-4o  & \texttt{Llama3.2-8B} & 2.49 (\textbf{↑0.90}) & 2.84 (\textbf{↑1.29}) \\
%    Re-verified & \texttt{Llama3-8B} & 2.85 (\textbf{↑0.90}) & 2.51 (\textbf{↑0.95}) \\
%     & \texttt{Phi4-14B} & 3.16 (\textbf{↑0.87}) & 2.67 (\textbf{↑0.86}) \\
%     & \texttt{Llama3.1-8B}  & 3.09 (\textbf{↑0.81}) & 2.69 (\textbf{↑0.76}) \\
%     & \texttt{Gemma2-9B}  & 3.08 (\textbf{↑0.87}) & 2.90 (\textbf{↑0.90}) \\
%     & \texttt{Olmo2-13B} & 3.02 (\textbf{↑0.75}) & 2.47 (\textbf{↑0.27}) \\
%     & \texttt{FLAN-T5-XXL} & 2.94 (\textbf{↑0.44}) & 3.02 (\textbf{↑0.37}) \\
%     \bottomrule
%     \end{tabular}
%     \caption{Caption}
%     \label{tab:my_label}
% \end{table}

\section{Study 2: Using the new norms dataset to predict human semantic judgments}

To assess whether the AI-enhanced norms in \texttt{NOVA} capture information about semantic structure beyond human-only norms or other approaches, we compared different approaches in their ability to predict human behavior in a triadic similarity judgment task \citep{jamieson2015next, hebart2023things, sievert2023efficiently}. In this task, participants must decide which of two option concepts is semantically more similar to a target concept. A candidate semantic embedding can ``predict'' human decisions by selecting whichever option word lies closer to the target word in the embedding space. We can assess the quality of the embedding by comparing how often the predicted response agrees with actual human decisions. In this study, we compared the predictions of \texttt{NOVA} embeddings to those based on the human-only feature norms and to those generated by a common word-embedding approach (\texttt{FastText}).   

We selected triplets designed to maximally discriminate \texttt{NOVA} and human-only feature norms. Thus for each trial, one of the option items was closer to the target in the human-only space while the other was closer in the AI-enhanced space (see Figure \ref{fig:part2_results}). We then computed how often the majority-vote across human participants agreed with the predictions of each embedding (AI-enhanced, human-only, FastText). If the AI-enhanced norms in \texttt{NOVA} contain information irrelevant to human-perceived semantics, their predictions should agree with human judgments less often than do those of the human-only norms. Furthermore, if either set of norms simply recapitulates the semantic structure evident in word embeddings, then predictions from the norms should do about as well as predictions from the FastText embeddings.  

\noindent\textbf{Generating maximally disagreeing triplets.}
To generate triplets that maximally differentiated the human-only and AI-enhanced norms, we computed cosine dissimilarity matrices for each set (Figure \ref{fig:heatmap_stats}B), Procrustes-aligned them to minimize disparity, and identified concepts with the largest discrepancies in their distances to other concepts. For example, in \texttt{NOVA} space, `accordion' was closer to `flute' than to `geyser', while the reverse was true in the human-only space (Figure \ref{fig:part2_results}A). We constructed 1,424 triplets where the two matrices produced divergent predictions, with each of the 786 concepts serving as the target approximately twice. The critical question was which matrix’s predictions would align more closely with human similarity judgments.

\noindent\textit{Participants}
31 participants were recruited from the UW-Madison psychology subject pool. Participants completed the task online for course credit. Each participant provided informed consent in compliance with the UW-Madison IRB.

\noindent\textit{Stimuli and Procedure}. The stimuli were the set of 1,424 triplets described above. Data were collected online via jsPsych \citep{de2015jspsych}. On each trial, a randomly selected triplet was displayed, with participants indicating which of two options was more similar to the target concept using a mouse click. All triplets were judged by each participant\footnote{there was data loss of a few trials for some participants due to technical issues.}.

% optimizing a measure of discriminability in the two spaces. Specifically, for any triplet of concepts with head $i$, left option $j$, and right option $k$:

% \begin{equation}
% \begin{split}
%     P_{human}(i,j,k) &= \frac{d_{ik}^{human}}{d_{ik}^{human} + d_{jk}^{human}}, \\
%     P_{llm}(i,j,k) &= \frac{d_{ik}^{llm}}{d_{ik}^{llm} + d_{jk}^{llm}}, \\
%     \text{Discrepancy}(i,j,k) &= -1 \times (P_{human}(i,j,k) - 0.5) \\
%     &\quad \times (P_{llm}(i,j,k) - 0.5),
% \end{split}
% \end{equation}

% where $d_{ik}$ represents the distance between concepts $i$ and $k$. Using this procedure we selected XXX triplets. All participants made judgments for all triplets.

\noindent\textbf{Results.}
Human similarity judgments agreed with predictions of the AI-enhanced norms for 86.20\% of triplets, a result unlikely to arise by chance ($p<$ 0.001, binomial test). Human judgments agreed with predictions of the FastText embeddings on 60.40\% of trials: reliably better than chance ($p<$ 0.001, binomial test), but significantly worse than the AI-enhanced embeddings (paired $t$-test, $t$(1,423) =18.37, $p<$ 0.001). Thus the richer structure evident in the AI-enhanced feature norms appears to better express human-discerned semantic similarity structure than to norms derived from humans alone or from word-embeddings.

\section{Discussion}

We presented a new approach for generating AI-enhanced semantic norms along with an accompanying dataset, \texttt{NOVA}.
We first conducted controlled experiments evaluating LLM feature verification performance against a reliable subset of human norm judgments, using the results to find an optimal model, prompting strategy, and verification strategy. We then applied the best-performing approach to generate an AI-enhanced large-scale norm dataset spanning over 750 concepts and over 8,000 features.
Concepts in the resulting NOVA dataset showed much higher feature densiry and a greater degree of feature overlap relative to the raw human-generated matrix. This overlap of features did not come at the cost of category selectivity, with concepts being reasonably organized into meaningful clusters. Finally, we used a triadic comparison task to show that NOVA vectors more accurately predicted human similarity judgments than did vectors based on human norms alone or word embeddings computed from natural language. The result suggests that AI-enhanced norms express semantic structure more similar to that discerned by human participants.

Taken together, our work addresses longstanding limitations in semantic norm generation by creating a dataset that includes a representative set of concepts and features, with AI-based feature verification validated against human jdugments. The feature density of the AI-enhanced norms reveals semantic similarity structure richer than previous norm datasets, unlocking the potential to better understand both the cognitive and the neural bases of semantic memory \citep{clarke2014object, rogersETAL2004, CoxETAL2024, fernandino2022decoding} and to guide the development of future computational neurocognitive models \citep{dilkina2008semantic,riordan2011redundancy, saxe2019mathematical, giallanzaETAL2024, suresh2024categories}. 
Lastly, the present work highlights the promise of integrating large-language models into workflows for cognitive science research in a controlled and verifiable manner and provides a replicable framework for future endeavors in this domain \citep{ suresh2023semantic, mukherjee2023human, trott2024can, dillion2023can, mukherjee2024large}.

% these enriched semantic norm datasets open exciting avenues for both theoretical and applied research. Beyond their traditional use in elucidating the structure of semantic memory \citep{ashcraft1978property, collins1975spreading}, these norms can be instrumental in refining computational models that predict semantic similarity \citep{mcrae2005semantic, dilkina2008semantic} and in mapping the neural substrates of conceptual processing \citep{clarke2014object, CoxETAL2024}. Moreover, by capturing a higher density and richer overlap of features, such datasets offer robust benchmarks for investigating semantic degradation in clinical populations \citep{creemcrae03, farahmcclelland1995,garrard2001longitudinal,rogersETAL2004} for probing the interplay between semantic representation and cognitive control \citep{giallanzaETAL2024}, and also for disambiguating different theories of how semantic representations interface with visual representations\citep{suresh2024categories}. In parallel, the integration of AI-generated responses not only enhances norm completeness but also paves the way for innovative applications of machine learning in cognitive science research \citep{trott2024can} and individualized assessments of conceptual structure.
\newpage
\section{Acknowledgments}
We thank members of the Knowledge and Concepts Lab at UW-Madison and the Neuroscience of Cognitive Control Lab at Princeton for helpful discussion and feedback.
This work was supported by a Vannevar Bush Faculty Fellowship (VBFF) administered through ONR to JDC and Multi University Research Initiative (MURI) award W911NF2110317 to TTR (co-PI).

\vspace{1em}
\fbox{\parbox[b][][c]{.45\textwidth}{\centering {All code and materials will be available at: \\
\href{https://github.com/Knowledge-and-Concepts-Lab/llm-norms-cogsci2025}{\url{https://github.com/Knowledge-and-Concepts-Lab/llm-norms-cogsci2025}}
}}}

\printbibliography

\end{document}